\documentclass{article}

\usepackage{microtype}
\usepackage{graphicx}
\usepackage{subcaption}
\usepackage{booktabs} 
\usepackage{tabularx}
\usepackage{array}
\usepackage{ragged2e}

\usepackage{hyperref}

\usepackage[preprint]{icml2026}
\usepackage{xurl}

\makeatletter
\renewcommand{\ICML@preprint}{}
\makeatother

\usepackage{amsmath}
\usepackage{amssymb}
\usepackage{mathtools}
\usepackage{amsthm}

\usepackage[capitalize,noabbrev]{cleveref}

\theoremstyle{plain}

\theoremstyle{definition}

\theoremstyle{remark}

\usepackage[textsize=tiny]{todonotes}

\icmltitlerunning{Uncensored Open-weight Models: Redistribution as the Persistence Layer}

\begin{document}

\twocolumn[
  \icmltitle{Uncensored Open-weight Models: \\ Redistribution as the Persistence Layer}



  \icmlsetsymbol{equal}{*}

  \begin{icmlauthorlist}
    \icmlauthor{10a Labs}{yyy}
  \end{icmlauthorlist}

  \icmlaffiliation{yyy}{Please cite this work as ``10a Labs (2026)". The full author list is available at the end of this report. Visit our website at \url{https://10alabs.com/} for more information} 
  

  \icmlcorrespondingauthor{Bobby McKenzie}{bobby@10alabs.com}

  \icmlkeywords{Machine Learning, ICML}

  \vskip 0.3in
]



\printAffiliationsAndNotice{}  

\begin{abstract}
A rapidly expanding ecosystem of actors is removing built-in safety guardrails from open-weight AI models. We profile this ecosystem by identifying key producers, downstream reproductions, and emerging applications. Between January 2024 and March 2026, we identified 3,471 original uncensored models on HuggingFace, each repackaged an average of 2.4 times; three actors account for 52\% of all 8,164 compressed redistributions. Once quantized and mirrored across separate accounts, formats, and registries such as Ollama, these models persist regardless of upstream removal and become easier to deploy downstream. Of the 1,643 identified GitHub applications integrating uncensored large language models (ULLMs), 25\% were classified as explicitly malicious.

\end{abstract}

\section{Introduction}

Major open-weight language model families are typically released with safety post-training intended to limit harmful behavior  \citep{gemmateam2026gemma4technicalreport, openai2025gptoss120bgptoss20bmodel, grattafiori2024llama3herdmodels, yang2025qwen3technicalreport, jiang2024mixtralexperts, Guo_2025}, but a rapidly expanding ecosystem of actors is removing these protections and redistributing modified versions—fueling downstream applications like offensive security tools, malware generators, and NSFW services—while the scale, key players, growth rate, and accessibility of this ecosystem remain poorly understood.

This report maps the full supply chain of safety-stripped or ``uncensored” models, from production on HuggingFace to deployment in GitHub applications—analyzing repositories from January 2024 to early March 2026 to assess the ecosystem’s scale and composition, key actors and their motivations, modification techniques and their evolution, acceleration dynamics and barriers to entry, and how model supply translates into downstream use cases and application demand. 

For our purposes, ``uncensoring” refers to techniques that intentionally strip the safety guardrails typically present when open-weight models are released to the public. These techniques target the refusal behavior trained into the model's weights, rather than external filters or classifiers. These techniques include activation-space abliteration \citep{arditi2024refusallanguagemodelsmediated, piras2026latentspaceattacksrefusalevasion}, which suppresses refusal directions in a model's internal representations, malicious fine-tuning \citep{halawi2024covertmaliciousfinetuningchallenges, qi2023finetuningalignedlanguagemodels}, such as refusal-filtered SFT, DPO \citep{rafailov2024directpreferenceoptimizationlanguage}, or LoRA \citep{hu2021loralowrankadaptationlarge} adapters that retrain the model to comply, and model merging, which blends uncensored models with other models to enhance capabilities like reasoning \citep{yang2025modelmergingllmsmllms, li2025modelmergingpretraininglarge}. When compressed and repackaged, the resulting models can be deployed locally on consumer hardware.

In total, we identified 3,471 original uncensored models on HuggingFace. Our key findings include:

\begin{itemize}
    \item{\textbf{Redistribution drives persistence.} Each original uncensored model is repackaged an average of 2.4 times. Three redistributors on HuggingFace (mradermacher, Triangle104, RichardErkhov) account for 52\% of all 8,164 compressed redistributions. A single producer’s 192 originals (huihui-ai) generated ~1,800 downstream repacks — roughly 14\% of the entire dataset — through other actors.}
    \item{\textbf{The producer and redistributor tiers are functionally separate.} At least 1,055 producers and at least 1,011 redistributors operate mostly as distinct populations; only 24\% of producers also redistribute. Redistributors source from public producer repositories without observed explicit coordination, and some operate request-driven quantization pipelines that respond to downstream demand rather than upstream release.}
    \item{\textbf{Deployment tracks accessibility, not raw supply.} GitHub application creation rose from ~30 per month in mid-2024 to 140–188 per month by late 2025. The surge began accelerating in mid-2025, aligned with maturation of the Ollama distribution layer, rather than with the later Heretic-driven \citep{heretic} surge that started in November 2025. 43\% of applications reference Ollama in their READMEs; 14\% reference direct HuggingFace downloads.}
    \item{\textbf{A small set of model families carries disproportionate downstream weight.} The Dolphin \citep{dolphin} family alone powers 30\% of the 1,643 identified applications. Ten actors account for 45\% of all non-dataset HuggingFace repositories. Concentration at the application layer reflects registry accessibility as much as model choice.}
    \item{\textbf{Chinese-origin foundation models sit at the center of the source-model landscape.} Chinese foundation models account for 38\% of all identified uncensored model repositories. Their share of new uncensored production rose from 1\% in Q1 2024 to 55\% in Q2 2025 and has since fluctuated near parity with Western-origin models, driven primarily by Alibaba’s Qwen family \citep{yang2025qwen3technicalreport}.}
    \item{\textbf{Commercial services participate in the same supply chain.} Venice.ai commissioned Dolphin-Mistral-24B-Venice-Edition from the Dolphin family’s creator, which offers uncensored products at tiered pricing, and appears in 77 GitHub repositories that integrate either its API or its Dolphin-Venice model. The commissioned model spawned 73 HuggingFace redistributions by community actors — placing a commercial release inside the same redistribution circuitry that carries hobbyist output.}
\end{itemize}

\begin{table*}[t]
\centering
\caption{Top redistributors by share of all observed compressed redistributions.}
\label{tab:top-redistributors}
\small
\setlength{\tabcolsep}{5pt}
\renewcommand{\arraystretch}{1.15}

\begin{tabularx}{\textwidth}{
    >{\RaggedRight\arraybackslash}p{0.15\textwidth}
    >{\RaggedRight\arraybackslash}p{0.10\textwidth}
    >{\RaggedRight\arraybackslash}p{0.10\textwidth}
    >{\RaggedRight\arraybackslash}p{0.09\textwidth}
    >{\RaggedRight\arraybackslash}X
}
\toprule
\textbf{Actor} &
\textbf{Redists.} &
\textbf{Share of observed total} &
\textbf{Selectivity} &
\textbf{Notes} \\
\midrule

\texttt{mradermacher} &
2,905 &
36\% &
80\% &
Active Jan.\ 2024--Mar.\ 2026; public request system
(${\sim}3{,}000$ discussions); semi-automated; self-describes
as a team. \\

\texttt{Triangle104/*} &
936 &
11\% &
74\% &
Selects models using uncensoring-related keywords in repository
metadata; profile and models have been removed from Hugging Face
since data collection. \\

\texttt{RichardErkhov} &
364 &
4\% &
6\% &
Student developer in Cyprus; affiliated with ``Team
mradermacher''; uncensored content is incidental to more than
26,000 total models. \\

\texttt{mlx-community} &
218 &
3\% &
45\% &
Community of approximately 4,000 members that converts models
for Apple Silicon. \\

\texttt{tensorblock} &
127 &
2\% &
Low &
Three-person automated GGUF conversion service. \\

\bottomrule
\end{tabularx}
\end{table*}

\section{Methodology}
\subsection{Overview}
We scraped HuggingFace using keyword searches across its model and dataset registries (including Chinese and Japanese terms) and crawls of all repositories published by known prolific producers. The keyword list included 43 high-signal terms plus 12 context-dependent terms, including Chinese and Japanese equivalents. Our repository search also included exhaustive crawls of 15 known prolific producers and dataset-registry search. This search surfaced 17,727 candidate repositories. Unless otherwise stated, counts in this report refer to HuggingFace repositories rather than unique sets of model weights.

An LLM classifier, GPT-5 \citep{singh2026openaigpt5card}, categorized each of the 17,727 candidate repositories as an original uncensored model, compressed redistribution, model merge, malicious dataset, or false positive, producing 12,360 HuggingFace repositories associated with safety guardrail removal. These included 3,471 original uncensored models, 8,164 compressed redistributions, 547 model merges, and 178 malicious datasets. All temporal analysis in this report is based on each repository's HuggingFace publication date, not the date of collection.

We also scraped GitHub repositories to find references to uncensored models. We identified 44,705 candidate repositories through model-name keyword search (59 terms), application-level keyword search (37 terms), 21 topic tags, actor crawls, and supplemental Chinese and Japanese keyword searches. After classification and filtering to applications only, 1,643 repositories were identified as integrating, recommending, or defaulting to an uncensored model backend.

\subsection{Producers and Redistributors}

We define two functionally distinct tiers of the uncensored model ecosystem: producers who remove safety behavior from open-weight models, and redistributors who repackage the results into formats deployable on consumer hardware. We classify a repository as an original uncensored model when it represents a model whose safety behavior has been intentionally modified, including through abliteration, safety-removing fine-tuning, or related techniques. We classify a repository as a compressed redistribution when it primarily repackages an existing uncensored model into a deployment-oriented format, such as GGUF, AWQ, GPTQ, EXL2, or MLX, without performing the underlying uncensoring step. The redistribution tier is the source of the ecosystem’s persistence. Production is spread across a long tail of actors, while redistribution is concentrated and operationally centralized. That asymmetry, consisting of many originals, few repackagers, and repeated downstream copies, is what converts a scatter of individual releases into a durable, enforcement-resistant supply.

\section{Results}
\subsection{Redistribution Enables Durability}

\subsubsection{Two Tiers, Mostly Separate Populations}

We identified at least 1,055 producers and at least 1,011 redistributors active on HuggingFace within the collection window. The two populations overlap by only 24\%. Redistributors source from producers’ public repositories without observed coordination, and in observed cases operate request-driven quantization pipelines that respond to downstream user demand rather than upstream release schedules.

\textbf{Production is long-tailed.} 61\% of producers published a single uncensored model. 22 actors account for 31\% of all originals. The top producers occupy distinct niches:
\begin{itemize}
    \item{Huihui-ai is weighted toward Chinese-origin base models (61\% of its 192 originals).} 
    \item{MuXodious covers 72 unique base-model families in 79 days using the Heretic tool.}
    \item{SicariusSicariiStuff has operated on refusal-filtered fine-tuning since January 2024, predating Heretic by nearly two years.}
\end{itemize}

\textbf{Redistribution is concentrated.} Three actors account for 52\% of all 8,164 compressed redistributions. The concentration holds across role types: 
\begin{itemize}
\item{Mradermacher operates a public model-request system with roughly 3,000 community discussions and self-describes as a team.}
\item{Triangle104 selects models whose names or metadata contain uncensoring keywords.}
\item{RichardErkhov self-identifies as “Team mradermacher” and routes uncensored content incidentally within a much larger conversion pipeline.}
\end{itemize}

Table \ref{tab:top-redistributors} reports each actor’s redistribution count, share of the observed total of all 8,164 compressed redistributions, and selectivity, defined here as the percentage of the actor’s total output that consists of uncensored models.

\subsubsection{The 2.4x Repackaging Effect on Distribution}

Each original uncensored model is repackaged an average of 2.4 times into compressed, quantization-optimized formats (GGUF, AWQ, GPTQ, EXL2, MLX) by downstream actors. The multiplier is unevenly distributed: huihui-ai’s 192 originals have been repackaged into roughly 1,800 compressed redistributions by other actors, accounting for approximately 14\% of the entire HuggingFace dataset identified in this collection.

The operational consequence is structural persistence. Availability of a given model no longer depends on its original producer. Enforcement action against a single upstream repository removes one node in a mirrored distribution graph rather than the model itself. Repackaged copies persist across separate accounts, separate formats optimized for different runtimes, and separate registries — principally HuggingFace and Ollama — that are not centrally controlled by any one provider.

This pattern also appears in the GitHub application data: application-layer creation responds more strongly to deployability signals (quantized formats, Ollama registrations) than to raw upload volume on HuggingFace. A model that exists on HuggingFace but is not quantized and mirrored is not, in practical terms, available at scale to application developers. The redistribution tier is the step that converts a model from “published” into “deployable.”

\subsection{Lower Barriers Expand the Ecosystem}

Redistribution persists because upstream production remains both scalable and varied. Automation expanded the volume of uncensored model production, while the available pool of source models broadened across Western- and Chinese-origin foundation model families. Together, those shifts sustained the flow of new models into the redistribution tier.

\begin{figure*}
    \centering
    \includegraphics[width=0.7\linewidth]{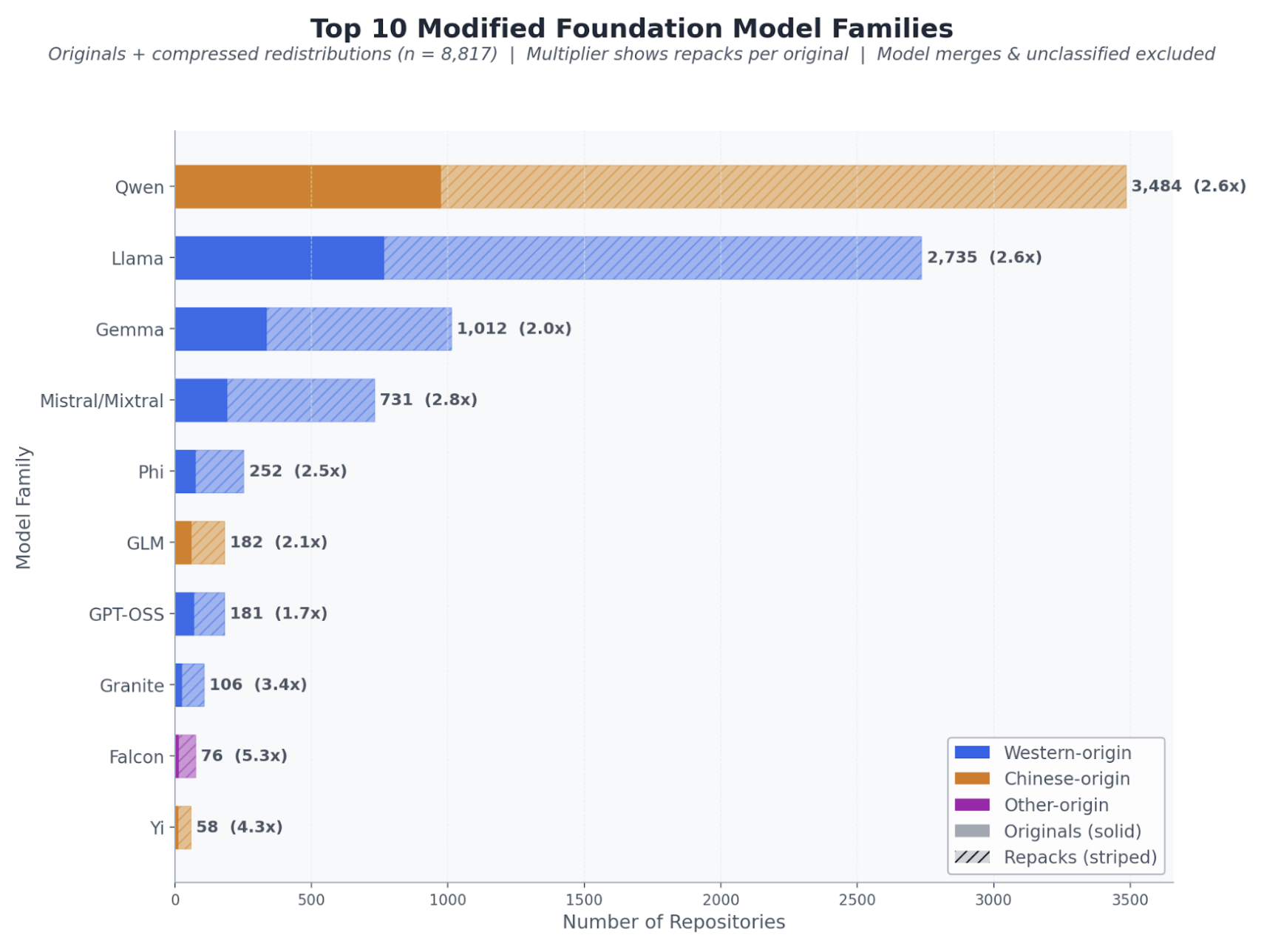}
    \caption{Top 10 modified foundation model families. Solid bars show originals; striped bars show compressed redistributions. Multiplier indicates compressed redistributions per original. Color indicates origin: blue = Western lab (Meta, Google, Mistral, etc.), gold = Chinese lab (Alibaba, DeepSeek, Zhipu, etc.). Model merges (4.4\% of the ecosystem) were excluded; merges most commonly involve Llama and Mistral base models.
}
    \label{fig:families}
\end{figure*}

\subsubsection{Heretic as the Production-Side Accelerant}
Before November 2025, uncensoring required working knowledge of transformer internals — loading model weights, identifying refusal directions in activation space, and modifying them through a Python workflow built around TransformerLens or refusal-filtered supervised fine-tuning. Heretic CLI \citep{heretic}, released on GitHub in late 2025, reduced that workflow to a single terminal command.

The transition was near-immediate. Zero Heretic entries appeared in October 2025. In November, 132 originals were produced using the tool, representing 55.5\% of that month’s output. By Q1 2026, Heretic accounted for 54\% of new original uncensored model production. In the 22 months before Heretic’s release, producers created approximately 89 original models per month. In the 4.5 months after, the rate rose to approximately 338 per month.

Heretic also expanded the producer base from roughly 640 to over 1,055 actors. Heretic users produce an average of 4.6 models each (compared to 3.5 for TransformerLens and 2.1 for refusal-filtered supervised fine-tuning) and have the lowest one-shot abandonment rate of any technique at 48.7\%, versus 69–75\% for fine-tuning methods. The tool expanded the pool of repeat producers rather than only adding one-off contributors — meaning the supply feeding the redistribution tier grew in both output volume and durability.

\subsubsection{Foundation Model Families Shape the Source-Model Landscape}

Figure \ref{fig:families} shows the top ten foundation model families for modified models on HuggingFace. 57\% of such models derive from Western-origin foundation models, 38\% from Chinese-origin models, 1\% from models based in other countries (e.g., UAE and South Korea), and 4\% from models whose origin could not be identified from available metadata. Three Western families – Meta’s Llama (52\%) \citep{grattafiori2024llama3herdmodels}, Google’s Gemma (17\%) \citep{gemmateam2026gemma4technicalreport}, and Mistral AI’s Mistral/Mixtral (15\%) \citep{jiang2024mixtralexperts} – account for 84\% of Western-origin entries where the base model could be identified. Among Chinese-origin models, Alibaba's Qwen family \citep{yang2025qwen3technicalreport} accounts for 80\%. Producers concentrate on the 3–8 billion parameter range (41\% of all entries), the size range most readily deployed on consumer-grade GPUs and laptops.

\subsubsection{Chinese-Origin Foundation Models as Increasingly Common Source Material}

\begin{figure}
    \centering
    \includegraphics[width=\linewidth]{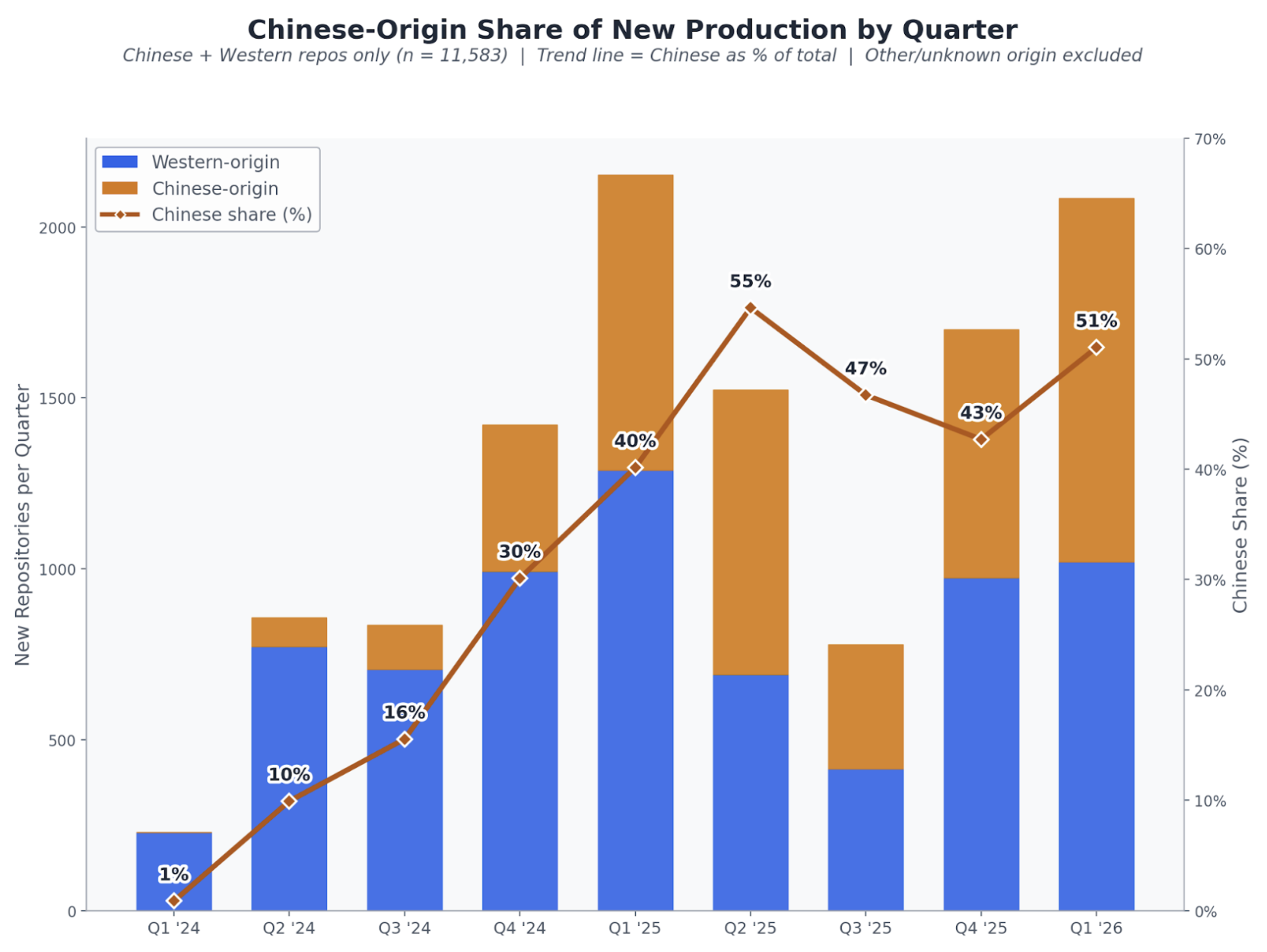}
    \caption{Chinese-Origin Share of New HuggingFace Repositories by Quarter (models, redistributions, merges, and datasets). Stacked bars (left axis) show absolute Chinese and Western-origin repository counts; the trend line (right axis) shows the Chinese share as a percentage of the two. Repositories with unknown base model origin excluded (777 of 12,360).}
    \label{fig:chinese-origin-share}
\end{figure}

Figure \ref{fig:chinese-origin-share} shows the share of new HuggingFace repositories for models, redistributions, merges, and datasets of Chinese origin. In Q1 2024, Chinese foundation models accounted for 1\% of new uncensored model production in the dataset. By Q2 2025, their share had risen to 55\%. Since then, quarterly share has fluctuated between 43\% and 51\%, settling at near-parity with Western-origin base models. Across the full collection window, Chinese-origin models account for 38\% of all identified uncensored model repositories.

Alibaba’s Qwen \citep{yang2025qwen3technicalreport} family drives most of that shift, accounting for 80\% of identified Chinese-origin entries. DeepSeek \citep{Guo_2025} appears frequently as secondary source material, particularly in huihui-ai’s output. The shift reflects availability rather than actor preference: the most prolific Chinese-origin producer (huihui-ai) draws 61\% of its base-model selections from Chinese families, but other top producers remain predominantly Western-weighted (MuXodious 83\% Western; DavidAU 69\% Western). Actors appear to target Qwen primarily because it is a widely used open-weight family, not because they specifically shifted toward Chinese source material.

As Chinese-origin source models account for a larger share of uncensored repositories, the redistribution layer gains a broader pool of frontier-capable models to mirror. Chinese-origin models sit in the same redistribution circuitry as Western-origin models; mradermacher and Triangle104 repackage both without distinction.

\subsection{Downstream Deployment Pulls Selectively From the Mirrored Supply}

Downstream application development draws from the redistribution tier in a patterned way. A small number of model families and one registry in particular (Ollama) carry most of the observed deployment weight. The downstream layer does not consume the full breadth of available uncensored models; it consumes the portion that has been made easy to integrate.

\subsubsection{Application Landscape and Intent}

\begin{figure*}
    \centering
    \includegraphics[width=0.7\linewidth]{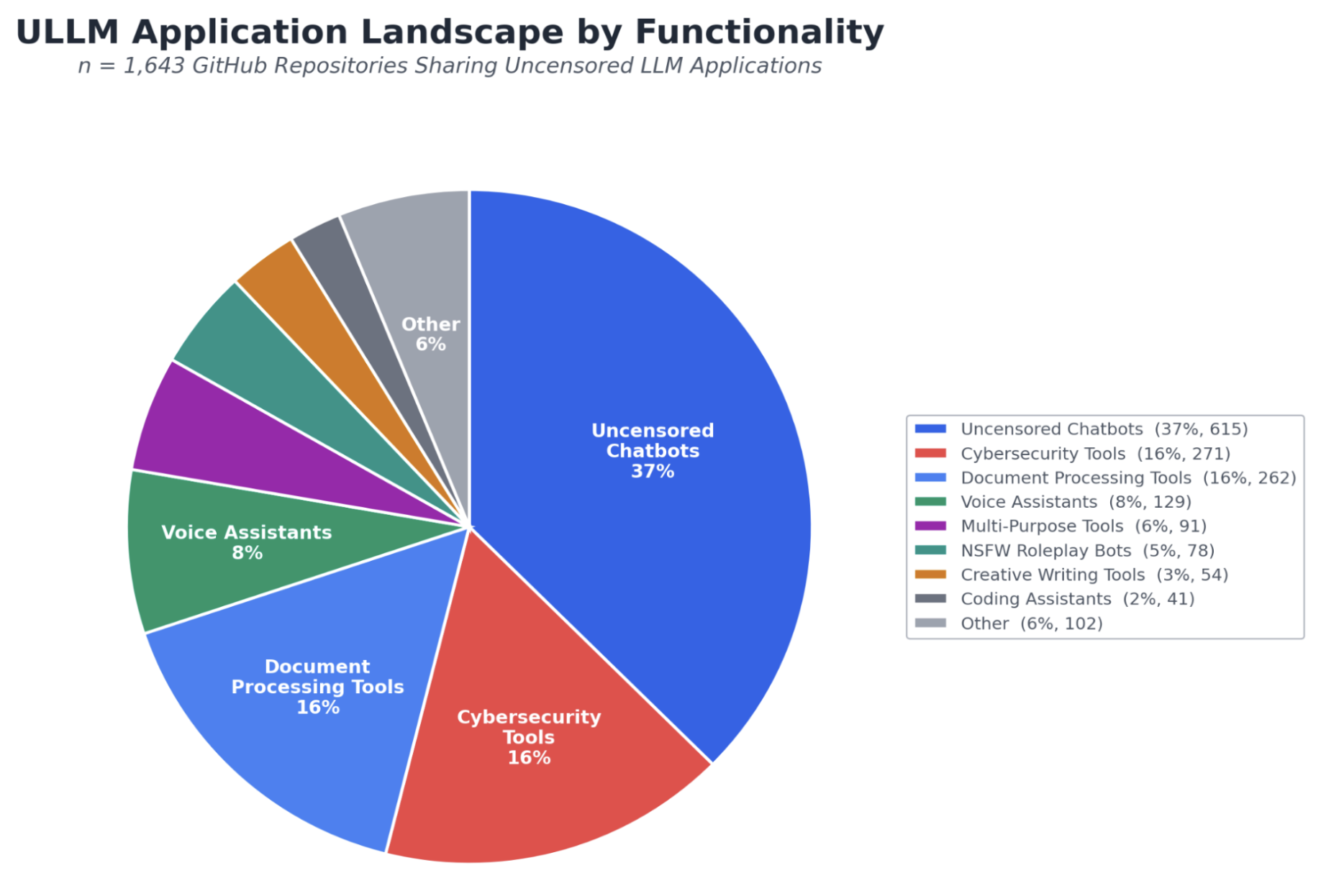}
    \caption{ULLM Application Landscape by Functionality Type. Uncensored chatbots dominate at 37\%, followed by cybersecurity  and document processing tools (16\% each). n = 1,643 GitHub repositories.}
    \label{fig:functionality-type}
\end{figure*}

We classify uncensored LLM applications by functionality type in Figure \ref{fig:functionality-type}. We identified 1,643 GitHub repositories that integrate, recommend, or default to an uncensored model backend. These uncensored large language model (ULLM) applications collectively account for over 11,600 stars and 1,800 forks. The categories break down as follows: general-purpose uncensored chatbots (37\%), cybersecurity tools (16\%), document processing pipelines (16\%), with the remainder spanning NSFW roleplay, NSFW storytelling, coding-related tools, and other specialized use cases.

We classified 411 applications (25\%) as explicitly malicious — designed for hacking, fraud, malware generation, or content that specifically exploits the absence of safety guardrails. The remaining 1,232 (75\%) were classified as ambiguous, meaning the functionality could serve both legitimate and harmful purposes. However, the removal of safety constraints is a deliberate design choice rather than an incidental property.

\subsubsection{Concentration at the Application Layer Mirrors the Supply Concentration}

\begin{figure*}
    \centering
    \includegraphics[width=\linewidth]{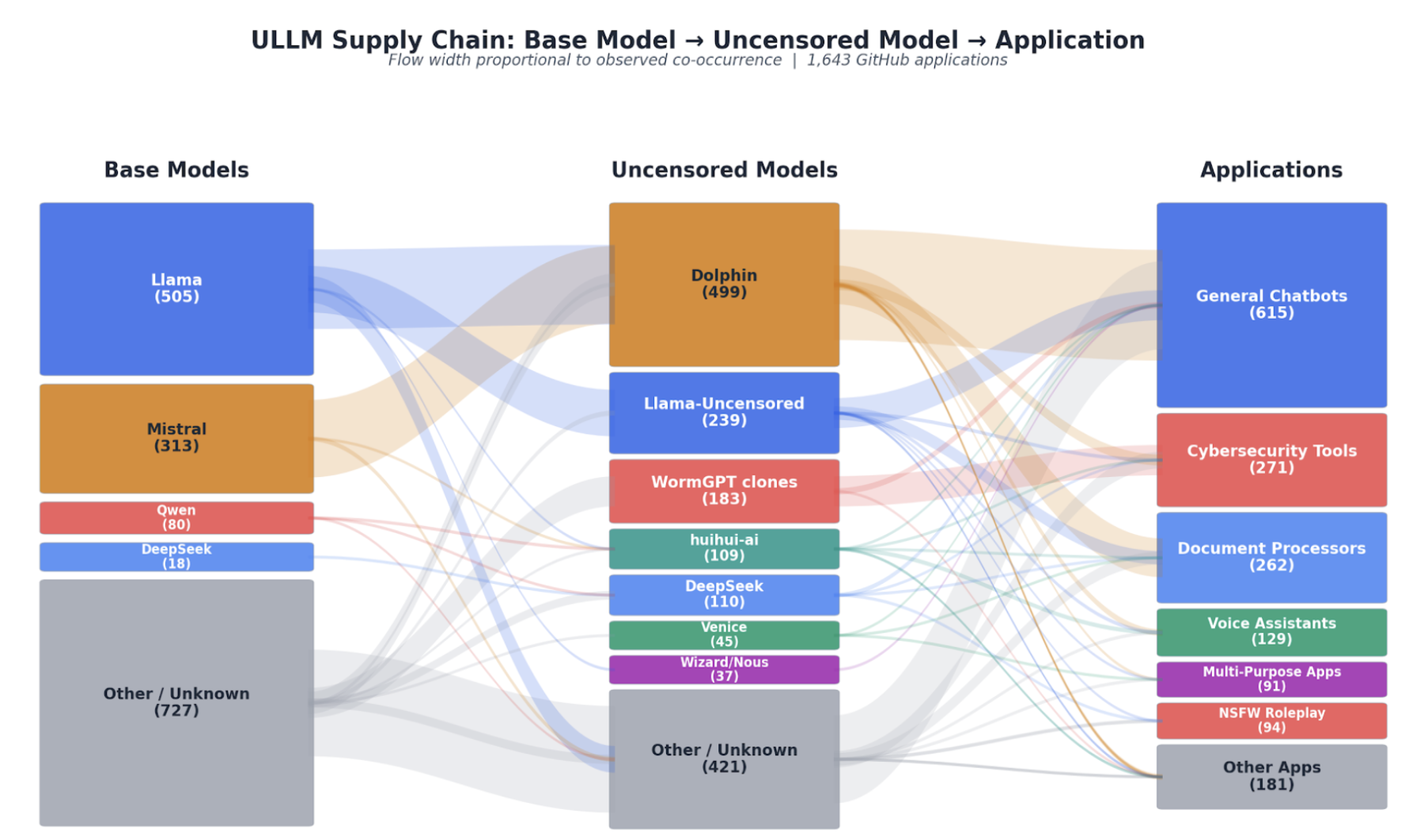}
    \caption{ULLM Supply Chain flow. The left column shows base foundation models; the middle column shows uncensored model families built from them; the right column shows downstream GitHub application categories. Flow width is proportional to observed co-occurrence in the dataset. n = 1,643 GitHub applications.}
    \label{fig:supply-chain-flow}
\end{figure*}

We map the supply chain flow of uncensored LLMs in Figure \ref{fig:supply-chain-flow}. The Dolphin family alone powers 30\% of identified applications (499 repositories). Dolphin’s dominance is driven primarily by its default availability in the Ollama model registry, which reduces integration to a single command, and secondarily by its creator’s prolific output. The creator has published 1,566 HuggingFace repositories, including compressed redistributions under the dphn account. This distribution advantage is reflected in adoption patterns: Ollama appears in 43\% of application READMEs, compared to 14\% for direct HuggingFace downloads.

Huihui-ai models power fewer applications (109) but command the highest aggregate star count of any identified-backend cluster (4,264 stars). This suggests huihui-ai models are favored in higher-visibility projects, particularly Chinese-language uncensored chat applications, even though Dolphin carries more raw integration volume.

The asymmetry between upload volume and deployment volume shows that downstream application growth depends more on model accessibility than on total model supply. HuggingFace uploads peaked at 879–1,077 models per month in early 2026. GitHub application creation rose from approximately 30 per month in mid-2024 to 140–188 per month by late 2025. The acceleration began in mid-2025—preceding the Heretic-driven production surge—and instead coincided with the maturation of the Ollama distribution layer. Application growth aligns more closely with model deployability and registry accessibility than with growth in raw model supply.

\subsubsection{Telegram and Secondary Channel Integration}

GitHub is itself the primary distribution channel for application code, but deployment extends to other platforms. 110 application repositories reference Telegram integration, deploying uncensored models as Telegram bots across uncensored chat, NSFW roleplay, and cybersecurity tooling. These bots sit downstream of the HuggingFace-to-Ollama-to-GitHub pipeline; they consume quantized models through the same redistribution circuitry but expose them through conversational interfaces to non-technical end users.

\subsection{Case Studies: The Pipeline in Practice}

Three cases illustrate how the production–redistribution–deployment pipeline behaves at different points in the supply chain: (1) a commercial service that commissions upstream and distributes downstream; (2) a single-actor offensive security deployment; and (3) a fragmented consumer-facing segment that consumes the redistribution layer without contributing to it.

\subsubsection{Venice.ai: Commercial Participant in the Same Circuitry}

Venice.ai \citep{venice} represents the most commercially developed deployment identified in the paired HuggingFace and GitHub collection. The company operates a full-stack AI API marketplace offering 144 models across text, image, video, and audio, including mainstream models (GPT-5.4, Claude Sonnet 4.6, Gemini) alongside explicitly uncensored offerings.

Venice directly commissioned the creator of the Dolphin family to produce Dolphin-Mistral-24B-Venice-Edition from Mistral-Small-24B using specialized fine-tuning and orthogonalization techniques, reportedly achieving a 2.2\% censorship refusal rate \citep{venice-comm-dolphin}. The commissioned model spawned 73 HuggingFace redistributions (GGUF, AWQ, MLX, EXL2) produced by the same community redistributors — mradermacher, bartowski, Triangle104 — identified in Section 3. On GitHub, 77 repositories integrate Venice’s API or its Dolphin-Venice model, spanning uncensored chatbots, cybersecurity tools, coding assistants, and at least one WormGPT clone.

Venice also offers dedicated uncensored products at tiered pricing: Venice Uncensored 1.1 (\$0.20/M input tokens), Venice Role Play Uncensored (\$0.50/M input), Lustify SDXL and Lustify v7 for NSFW image generation, and GLM 4.7 Flash Heretic — the last of which uses the same Heretic abliteration tool documented in Section 3.2.1. Venice therefore functions as both a consumer and a producer inside the uncensoring supply chain: it purchases upstream capacity from a hobbyist-origin producer, pushes the result into the same community redistribution layer, and then redistributes commercially through its own API to downstream developers.

\subsubsection{CyberAlbSecOP: Offensive Tooling on Top of a Redistribution Backbone}

GitHub user CyberAlbSecOP maintains two complementary tools: BLACKHATGOD Master Hacker GPT (201 stars) and HYDRAX Advanced Malware Generator GPT (25 stars), both powered by Dolphin 2.6 Mixtral. A single actor has built a coordinated offensive security suite on a model that reached them through the same Dolphin→Ollama→GitHub path traced elsewhere in the dataset.

The broader dataset contains 257 cybersecurity tool repositories and 9 dedicated malware generator repositories that sit in the same pattern. These are not novel models. They are conventional applications built on redistributed upstream weights, with the malicious configuration supplied by application-layer prompting, tool wiring, and UI design rather than by further model modification.

\subsubsection{NSFW Deployment: High Volume, Low Actor Coordination}

92 repositories are classified as NSFW roleplay (76) or NSFW storytelling (16). The highest-starred, NSFW-chatbot by samttoo22-MewCat (81 stars), uses ValueFX9507/Tifa-DeepsexV2-7b — a fine-tuned model purpose-built for explicit content. Others integrate Dolphin or DeepSeek directly, or instruct users to select “uncensored” models through Ollama at install time.

NSFW roleplay and storytelling repositories have low average star counts (3.7 per repository) but high volume, indicating a fragmented landscape of small-scale personal or niche-community projects rather than centralized services. This represents a segment of downstream demand that consumes the redistribution tier without contributing originals back to it — and that relies on Ollama’s registry model rather than direct HuggingFace integration for most deployments.

\section{Discussion}
\subsection{Implications for Observability and Enforcement}

The data point to three structural features that shape availability, concentration, and observability in the uncensored model ecosystem. Together, they indicate where supply persists, where activity concentrates, and where downstream use is most visible.

\textbf{Removal at the producer tier does not meaningfully reduce availability.} The 2.4x redistribution multiplier means that, in aggregate, original models generate multiple downstream copies. Takedowns at the producer level remove upstream authorship but not downstream supply. Triangle104’s profile and models, for example, were removed from HuggingFace after the collection window; the 936 compressed redistributions published by that account were produced prior to its removal and continue to persist as downstream copies across the ecosystem.

\textbf{The redistribution tier is the ecosystem’s narrowest point.} Three actors produce 52\% of all compressed redistributions. The producer tier is long-tailed and diffuse, while the redistribution tier is concentrated. That concentration means changes at a small number of redistribution nodes would affect a larger share of observed downstream availability than comparable action at the producer tier. The concentration reflects operational capacity — semi-automated pipelines, request systems, and team structure — rather than explicit coordination.

\textbf{Deployment observability depends on the application layer, not the model layer.} 25\% of identified applications are explicitly malicious by design, and the ambiguous 75\% deliberately integrate uncensored backends. Models themselves disclose relatively little about downstream use. Application-layer metadata — README references, registry sources, functionality category — carries more signal about actual deployment patterns than model-layer metadata does.

\textbf{Commercial and hobbyist deployment rely on the same distribution layer.} Venice.ai’s participation demonstrates that commercial services and community-driven projects move through the same redistribution layer. Observability at that layer therefore captures both segments at once, but it also means changes affecting that layer would reach both segments regardless of intent.

\subsection{Limitations}
We acknowledge a few limitations of this work:
\begin{itemize}
\item{\textbf{Platform scope.} HuggingFace for model supply; public GitHub for application deployment. ModelScope, Gitee, private repositories, self-hosted deployments, and direct-download channels are not captured.}
\item{\textbf{Classification validation.} Single LLM classifier pass without human ground-truth validation. 30.3\% of retrieved candidates were classified as false-positive search results. We did not measure precision or recall.}
\item{\textbf{Technique attribution gap.} 21\% of originals carry no technique signal in metadata; a further 17\% are identified as abliteration at the category level but lack a specific tool attribution.}
\item{\textbf{Partial periods.} Q1 2026 covers approximately 2.3 months (January through early March) and is not directly comparable to complete quarters.}
\item{\textbf{GitHub search ceiling.} The GitHub Search API caps results at 1,000 per query. High-frequency CJK terms are truncated; the true population of non-English applications is likely larger than reported.}
\item{\textbf{Origin resolution.} 6.3\% of HuggingFace entries have unresolved base-model origin, primarily repacks of custom-named models lacking base model metadata.}
\end{itemize}

\section{Related Work}

A few works have investigated the uncensored model ecosystem. \citet{sokhansanj2025uncensoredaiinthewild} profiles 8,608 uncensored models from HuggingFace repositories. They analyze refusal rates on safety tasks and find that uncensored models comply with 74.1\% of unsafe requests. They also note that the top 5\% of producers they identify account for 60\% of the uncensored models in their study. \citet{lin2025consiglieresshadowunderstandinguse} investigate the landscape of uncensored models through the lens of cyber crime. They represent the ecosystem as a knowledge graph and use graph-based deep learning to identify 11,000 uncensored models based on a small dataset. Other works such as \cite{malla} shift focus from public repositories on HuggingFace and GitHub to underground marketplaces, identifying techniques that malicious actors in these communities use to remove safeguards from open-weight models. 

\section{Conclusion}

Safety post-training is an important mechanism for limiting harmful behavior in many open-weight models. A growing ecosystem of developers has used sophisticated techniques to remove these guardrails from models and diffuse uncensored versions across popular platforms such as HuggingFace and GitHub. In this work, we identify and profile model producers and redistributors active in this ecosystem. We track 3,471 original uncensored model repositories and 8,164 redistributions, noting key users who are responsible for an outsized portion of uncensored model activity. We further identify open-source applications that employ these models, including malware generation tools and NSFW content generators. We hope this work inspires further research into the ecosystem of uncensored models.

\section*{Disclaimer}

This report does not contain non-public personally identifiable information (PII). All data is sourced from publicly accessible platforms, and any included user handles or identifiers reflect publicly visible information only. For this report, we did not conduct attribution, identity resolution, or deanonymization of any actors referenced.

\section*{Author List}
Please cite this report as ``10a Labs (2026)". The complete list of authors is presented in alphabetical order. All authors were affiliated with 10a Labs during this project.

Juliette Garcia, Hailey May, Bobby McKenzie, David Pham, Matthew Swain, Joshua Valdez, Corie Wieland, Zachary Yahn

\bibliography{sources}
\bibliographystyle{icml2026}

\newpage
\appendix
\onecolumn


\end{document}